\pdfoutput=1

\documentclass[11pt]{article}

\usepackage[]{ACL2023}

\usepackage{times}
\usepackage{latexsym}

\usepackage[T1]{fontenc}

\usepackage[utf8]{inputenc}

\usepackage{microtype}

\usepackage{inconsolata}
\usepackage{amsmath}
\usepackage{xcolor}
\usepackage{dirtytalk}
\usepackage{booktabs}
\usepackage{subfig}
\usepackage{url}
\usepackage{graphicx}
\usepackage{amssymb}
\usepackage{color}
\usepackage[normalem]{ulem}
\useunder{\uline}{\ul}{}

\definecolor{ForestGreen}{RGB}{34,139,34}

\title{Laying Anchors: Semantically Priming Numerals in Language Modeling}

\author{Mandar Sharma \\
  Virginia Tech \\
  \texttt{\normalsize mandarsharma@vt.edu} \\\And
  Rutuja Murlidhar Taware \\
  Virginia Tech \\
  \texttt{\normalsize trutujamurlidhar@vt.edu} \\\AND
  Pravesh Koirala \\
  Vanderbilt University \\
  \texttt{\normalsize pravesh.koirala@vanderbilt.edu} \\\And
  Nikhil Muralidhar \\
  Stevens Institute of Technology\\
  \texttt{\normalsize nmurali1@stevens.edu} \\\And
  Naren Ramakrishnan \\
  Virginia Tech \\
  \texttt{\normalsize naren@cs.vt.edu} \\}

\begin{document}
\maketitle

\begin{abstract}
Off-the-shelf pre-trained language models have become the de facto standard in NLP pipelines for a multitude of downstream tasks. However, the inability of these models to properly encode numerals limits their performance on tasks requiring numeric comprehension. We introduce strategies to semantically prime numerals in any corpus by generating anchors governed by the distribution of numerals in said corpus, thereby enabling mathematically grounded representations of these numeral tokens. We establish the superiority of our proposed techniques through evaluation on a range of numeracy tasks for both in-domain (seen) and out-domain (unseen) numerals. Further, we expand our empirical evaluations to numerals ranging from 1 to 10 billion, a significantly broader range compared to previous studies of the same nature, and we demonstrate significant improvements in the mathematical grounding of our learned embeddings.\footnote{Our codebase with the data and pre-trained models are hosted at \url{https://github.com/Mandar-Sharma/Laying-Anchors}}
\end{abstract}

\section{Introduction}
\begin{figure}[!ht]
    \centering
    \includegraphics[scale=0.3]{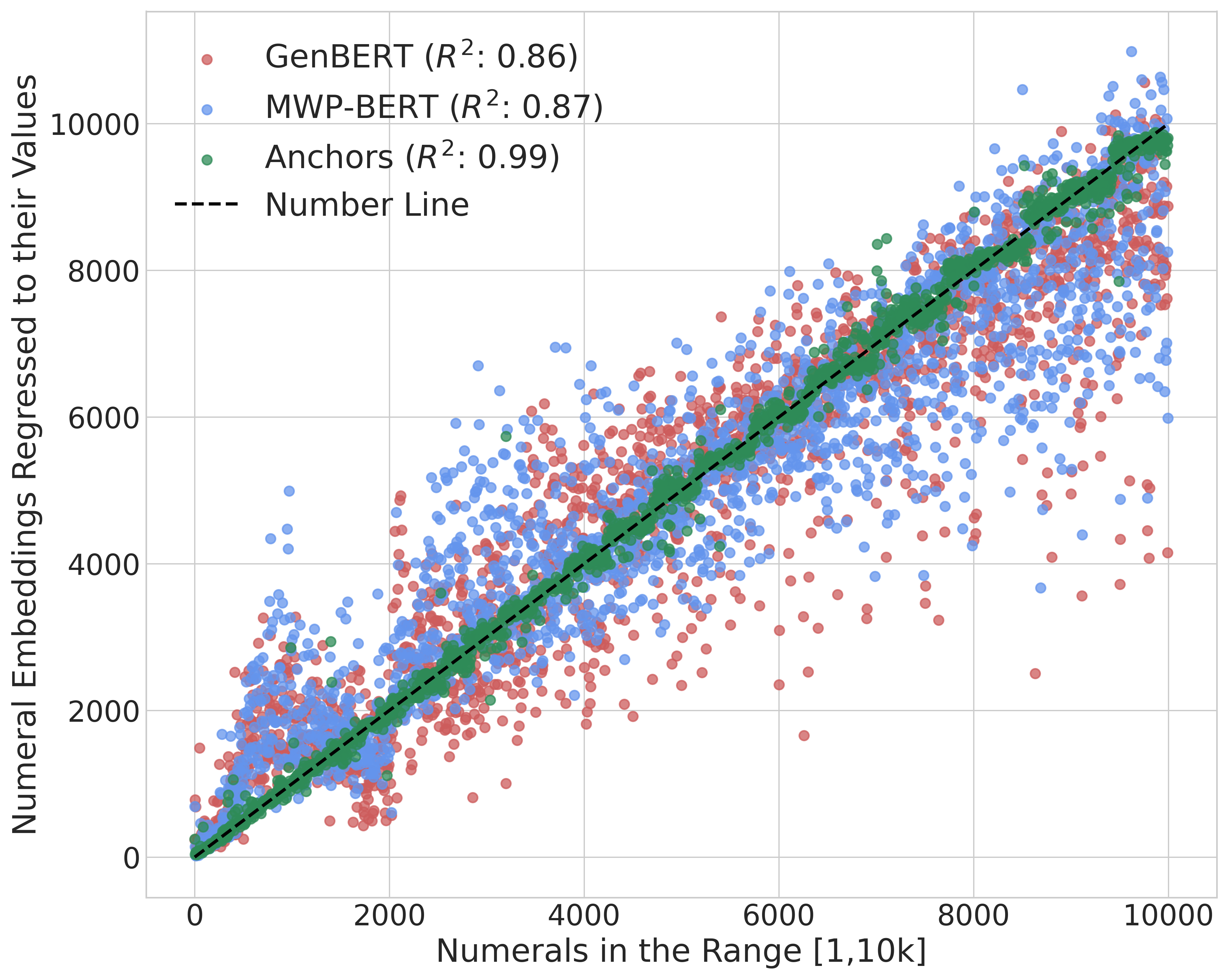}
    \caption{\footnotesize \textit{Anchor-based embeddings correlate significantly better to the number line:} The plot above showcases how well the numeral embeddings from the baselines and our model (Anchors) correlate to the number line with their $R^{2}$ goodness-of-fit scores presented. The numeral range {[1,10k]} is employed for this plot as it contains a healthy mixture of both in-domain and out-domain numerals from our dataset.}%
    \vspace{-0.5cm}
    \label{fig:plots}
\end{figure}
\begin{figure*}[t]
    \centering
    \includegraphics[scale=0.2]{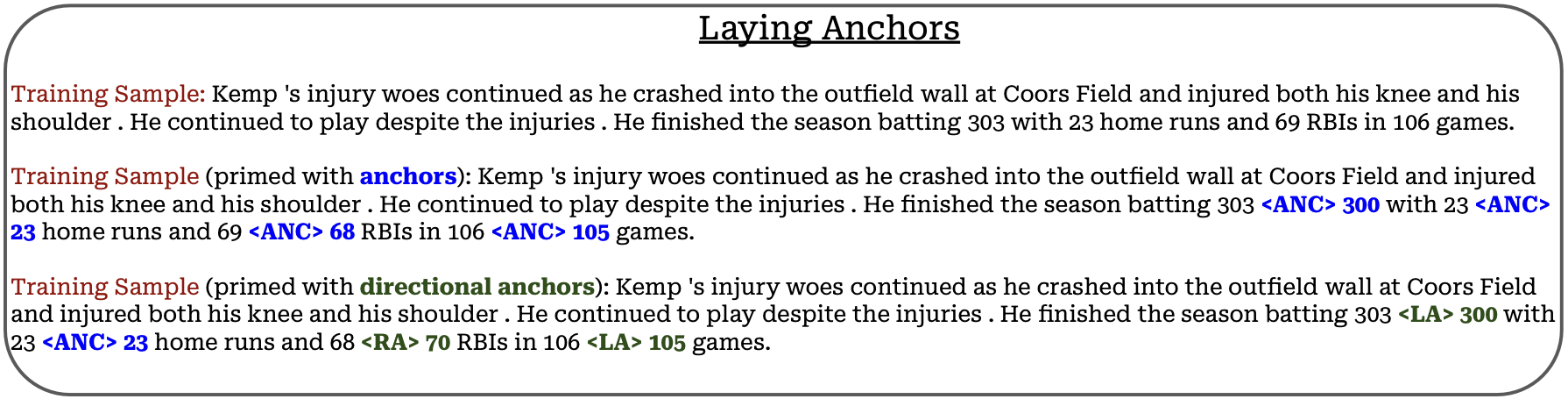}
    \caption{\footnotesize \textit{How are the numerals in the training corpus primed?} Showcasing samples from the training corpus - as-is, primed with simple anchors \scriptsize{\textbf{\textcolor{blue}{<ANC>}}} \footnotesize where each numeral in the sample is augmented with the its closest anchor, and directional anchors \scriptsize{\textbf{\textcolor{ForestGreen}{<LA>/<RA>}}} \footnotesize where the direction of the anchor with respect to the numeral (left or right in the number-line) is also embedded.}%
    \vspace{-0.5cm}
    \label{fig:anchors}
\end{figure*}

\textit{Numeracy}, at its core, is the comprehension of numbers, akin to the comprehension of words in literacy. The magnitude of a number is especially tied to its meaning \cite{dehaene:98}; as such, in developmental psychology, children able to distinguish numbers based on their magnitudes are said to possess the concept of numbers \cite{piaget:52}. In the context of NLP, because numbers often grant objectivity to language \cite{porter:96}, language models that can comprehend numeric magnitude and scales allow for better inference \cite{naik:18}, information extraction \cite{madaan:16}, and data-to-text generation \cite{sharma:2021, sharma:2022}. 

Numeric comprehension can indeed be induced in language models through explicit supervision \cite{vinyals:16}; however, the inherent numeric capabilities of off-the-shelf language models induced from unsupervised training have been shown to be inadequate \cite{naik:18} and often fail to extrapolate to numerals \textit{not seen} in the training set \cite{wallace:19, razeghi:22} - referred to as \textit{out-of-domain} (OOD) numerals. Approaches for numeracy induction to-date either involve strategies that learn representations for numerals separately from regular tokens \cite{spith:18, jiang:20} or do so by training models on numeracy-specific tasks \cite{mathnlp:20, liang:22}. In contrast, we \textit{prime} (see \textcolor{blue}{\textsection 2}) the numerals in the training corpus by laying anchors such that numeracy is induced via the unsupervised pre-training of the model itself without separately training numerical embeddings. As illustrated in Figure \ref{fig:plots}, our model shows substantial improvements in numeral representations for both numerals present in the training corpus (in-domain) as well as numerals absent from the training corpus (out-domain) over the state-of-the-art baselines. 

Further, the evaluation of numeracy in language models through their ability to predict numbers in a manner similar to textual tokens \cite{spith:18, chen:19} omits the influence of rote-memorization \cite{zhang:20}. In order to decouple the rote-memorization of numerals with respect to the linguistic context in which they appear, our study follows the evaluation protocols of \citet{wallace:19} wherein the quality of learned representations are assessed through a set of numeric comprehension tasks. Our contributions can be summarized as:
\begin{itemize}
    \item We develop new techniques for mathematical grounding of numerals in a corpus and quantitatively demonstrate significant improvements in model numeracy.
    \item We evaluate our models on a range of numerical tasks for numerals 1 to 10 billion ($10^{10}$), the largest analysis scope to the best of our knowledge, and evaluate its extrapolation capabilities to unseen (out-domain) numerals.
    \item Through rigorous evaluation, we demonstrate that the anchoring mechanisms lead to improved magnitude estimation (from \textit{compressive representations}) and relative ordering (from \textit{directional priming}) of numerals. 
\end{itemize}

\section{Priming Numerals with Anchors}
\textbf{How does one prime numerals?} The \emph{priming} effect is a temporary change in the perception of a target stimulus that frequently occurs in conjunction with a priming stimulus \cite{bargh:00}. Similarly, semantic priming establishes the strength of relations among items belonging to the same or different categories \cite{zorzi:04}. 

Now, what does this mean in the context of numerals in a training corpus? Consider numerals 0 and 10 that are both equidistant to a supposed anchor numeral 5. If a language model has never seen the numerals 0 and 10 in its training corpus, the anchor numeral 5---that the model has seen during its training---can now be used to ground the magnitudes of these unseen numerals such that the model can now reason its magnitude. \textit{Essentially, we intend to ground the magnitudes of numerals that the model rarely sees or has never seen based on the magnitudes of the numerals that it has frequently seen, known as the anchors}.
\vspace{0.25cm}

\noindent\textbf{How are the anchors determined?} First, we extract all numerals $X$ from a training corpus $C$ through which we intend to induce our anchors. The intuition that anchors should be numerals widely represented (frequent) in the corpus leads to the choice of Gaussian mixture models (GMMs) in contrast to clustering methods such as k-means that lack probabilistic cluster assignment. The set of anchors is induced from the means $\mu_{k}$ of each Gaussian $k \in K$ such that each numeral $n \in X$ can be tied to its closest anchor (\ref{gmm}). Here, $\mathcal{N}$ represents the probability density function and $\pi_{k}$,$\sigma_{k}$ represent the mixing coefficient and standard deviation for the $k$-th Gaussian component. The initialization and the choice of $K$ is described in \textcolor{blue}{\textsection A.1}.
\vspace{-0.2cm}
\begin{equation}
\label{gmm}
    p(n) = \sum_{k=1}^{K} \pi_{k} \mathcal{N} (n; \mu_{k}, \sigma_{k}^2)
    \vspace{-0.2cm}
\end{equation}

\noindent\textbf{Devising the four categories of anchors:} Theories for mental representation of cardinality further divides our implementation of these anchors into two halves: a continuous linear representation \cite{dehaene:03} and a compressive representation where the difference between numerals $n$ and $n+1$ decreases as $n$ increases \cite{dehaene:90}. As such, for linear representation of the number line, we associate numerals with their closest anchor without alteration - giving us our first model \textit{Anchors}. Similarly, for compressive representation, a given numeral $n$ is anchored to $m$ from a set of log-normalized anchors such that $\ln{(n)} \approx m$ - our second model \textit{ln Anchors}. In both these methods, the priming is implemented through a specialized token \textcolor{blue}{<ANC>} added to the tokenizer.

Further, this priming effect is known to be symmetric with respect to the priming direction and additive to the effect of repetition priming \cite{reynvoet:02}. This notion leads to our second category of models, viz. \textit{directional anchors} represented with bi-directional arrows $\rightleftarrows$. Thus, in addition to attaching anchors to numerals in the corpus, we signify \textit{where} the anchor lies in the number line with respect to the target numeral using specialized tokens \textcolor{ForestGreen}{<LA>} (stating the anchor lies to the left of the target numeral in the number line) and \textcolor{ForestGreen}{<RA>} (stating the anchor lies to the right of the target numeral in the number line). Training samples augmented with both \textcolor{blue}{<ANC>} and \textcolor{ForestGreen}{<LA>/<RA>} are depicted in Figure \ref{fig:anchors}.

\section{Experimentation and Results}
\vspace{-0.1cm}
As delineated in the previous section, we evaluate four configurations of our model pre-trained on the \textit{anchor-augmented} WikiText-103 corpus \cite{wikitext}: Anchors, \textit{ln} Anchors, Anchors ($\footnotesize{\rightleftarrows}$), and \textit{ln} Anchors ($\footnotesize{\rightleftarrows}$). The details of the datasets, pre-training and fine-tuning configurations, and embedding retrieval are described in \textcolor{blue}{\textsection A.2}.
\begin{table*}[t]
\centering
\scriptsize
\caption{\footnotesize \textit{For in-domain numerals, Anchors consistently showcases enhanced numeracy across all numeral ranges while the baselines suffer significant degradation for larger numeral ranges:} Performance of our model variants (Anchors) vs the baselines for in-domain numerals on four tasks evaluating the numeracy captured by model embeddings. The tasks are further sub-divided into number ranges and column $\forall$ Z $\in$ C includes all numerals $Z$ in corpus $C$.}
\begin{tabular}{@{}lcccccccccc@{}}
\toprule
Models     & \multicolumn{5}{c}{\textbf{Decoding} (Log-RMSE)}                                                   & \multicolumn{5}{c}{\textbf{Addition} (Log-RMSE)}                                                   \\ \midrule
Range      & {[}1,100{]}   & {[}100, 1k{]} & {[}1k, 10k{]} & {[}10k, $10^{10}${]} & $\forall$ Z $\in$ C                    & {[}1,100{]}   & {[}100, 1k{]} & {[}1k, 10k{]} & {[}10k, $10^{10}${]} & $\forall$ Z $\in$ C                    \\ \midrule
GenBERT    & 0.0926        & 0.0301        & 0.0215        & 0.0639           & 0.0700                 & 0.0250        & 0.0204        & 0.0237        & 0.0905           & 0.0752                 \\
MWP-BERT   & 0.0633        & 0.0213        & 0.0150        & 0.0540           & 0.0575                 & {\ul 0.0077}  & 0.0128        & 0.0200        & 0.0871           & 0.0533                 \\ \midrule
Anchors        & 0.1279        & 0.0196        & 0.0074        & 0.0344           & 0.0424                 & 0.0449        & 0.0172        & 0.0102        & 0.0442           & 0.0401                 \\
Anchors (\scriptsize{$\rightleftarrows$})     & 0.1269        & 0.0123        & 0.0057        & {\ul 0.0290}     & 0.0422                 & 0.0180        & 0.0122        & 0.0089        & {\ul 0.0426}     & 0.0378                 \\
\textit{ln} Anchors    & {\ul 0.0279}  & {\ul 0.0087}  & {\ul 0.0049}  & 0.0375           & {\ul \textbf{0.0304}}  & 0.0119        & {\ul 0.0067}  & {\ul 0.0084}  & 0.0572           & {\ul \textbf{0.0329}}  \\
\textit{ln} Anchors (\scriptsize{$\rightleftarrows$}) & 0.1729        & 0.0109        & 0.0054        & 0.0375           & 0.0525                 & 0.0157        & 0.0079        & 0.0106        & 0.0585           & 0.0443                 \\ \midrule
           & \multicolumn{5}{c}{\textbf{List Maximum} (Accuracy)}                                               & \multicolumn{5}{c}{\textbf{List Minimum} (Accuracy)}                                               \\ \midrule
           & {[}1,100{]}   & {[}100, 1k{]} & {[}1k, 10k{]} & {[}10k, $10^{10}${]} & $\forall$ Z $\in$ C                    & {[}1,100{]}   & {[}100, 1k{]} & {[}1k, 10k{]} & {[}10k, $10^{10}${]} & $\forall$ Z $\in$ C                    \\ \midrule
GenBERT    & 92.49\%       & 91.49\%       & 82.50\%       & 82.50\%          & 83.50\%                & 94.99\%       & 81.50\%       & 83.50\%       & 70.49\%          & 86.00\%                \\
MWP-BERT   & {\ul 93.00\%} & 91.50\%       & 85.00\%       & 79.00\%          & 87.25\%                & {\ul 96.00\%} & 88.50\%       & 88.50\%       & 75.00\%          & 87.00\%                \\ \midrule
Anchors        & 92.50\%       & 91.00\%       & 63.00\%       & 87.00\%          & 87.75\%                & 90.49\%       & 88.99\%       & 92.00\%       & 86.00\%          & 88.87\%                \\
Anchors (\scriptsize{$\rightleftarrows$})     & {\ul 93.00\%} & 83.00\%       & 82.50\%       & 83.00\%          & 88.37\%                & 92.50\%       & 90.00\%       & 86.50\%       & 85.50\%          & 91.00\%                \\
\textit{ln} Anchors    & 92.00\%       & 88.00\%       & 88.50\%       & 81.50\%          & 89.37\%                & 93.50\%       & 92.00\%       & 81.00\%       & 85.00\%          & 90.50\%                \\
\textit{ln} Anchors (\scriptsize{$\rightleftarrows$}) & 89.00\%       & {\ul 93.50\%} & {\ul 90.50\%} & {\ul 88.00\%}    & {\ul \textbf{89.87\%}} & 94.00\%       & {\ul 93.50\%} & {\ul 92.50\%} & {\ul 91.50\%}    & {\ul \textbf{92.50\%}} \\ \bottomrule
\end{tabular}
\label{table:1}
\end{table*}
\vspace{-0.2cm}
\subsection{Baselines}
\noindent\textbf{GenBERT} \cite{mathnlp:20}: This model is based on the pre-trained BERT model and is additionally trained for quantitative reasoning (arithmetic, list minimum/maximum operations) with a corpus of 1 million synthetically generated quantitative reasoning prompts.
\vspace{0.2cm}

\noindent\textbf{MWP-BERT} \cite{liang:22}: Also based on the pre-trained BERT model, MWP-BERT is trained for solving math word problmes (MWP) through the injection of numerical properties via multiple numeracy grounded pre-training objectives that encourages contextual representations to capture numerical information. 

\subsection{Numeracy of Embeddings}
In line with the premise set by \citet{wallace:19}, we evaluate the performance of the model embeddings on the tasks described below for different numerical ranges. The configurations for regressors and classifiers for the tasks mentioned below, are described in \textcolor{blue}{\textsection A.3}. 
\vspace{0.2cm}

\noindent\textbf{Decoding}: Given embeddings for a set of numerals, the task is to regress them to their numerical values, thus assessing the fidelity of the numerical magnitudes captured by the embeddings. 

\noindent\textbf{Addition}: Given sets of concatenated embeddings of two numerals, the task is to regress them to the numerical sum of the two numerals. In addition to assessing the magnitude fidelity, this task additionally requires number manipulation.  

\noindent\textbf{List Maximum-Minimum}: While the first two tasks assess the magnitude captured by the embeddings, the task of predicting the maximum or minimum numeral in a set of randomly sampled numerals assesses whether the embeddings capture relative ordering.

\section{Results}

 The results of above four tasks are illustrated in Table \ref{table:1} for in-domain numerals, and similarly in Table \ref{table:2} for out-of-domain numerals\footnote{Please note that as all numerals in range [1,100] and [100, 1k] appear in the training corpus, only numeral ranges [1k, 10k] and [10k, $10^{10}$] qualify for OOD evaluation.} (see \textcolor{blue}{\textsection A.3}). Our findings paint a consistent picture:
 \begin{itemize}
     \item For the lower numeral ranges $[1,100]$ and $[100,1k]$, all models do seemingly well. However, the performance of the baselines decreases sharply as the magnitude of numerals increase (for ranges $[1k,10k]$ and $[10k, 10^{10}$]). However, \emph{Anchors} and its variants have consistent performance across all the numeral ranges for both in-domain numerals and out-of-domain numerals.
     \item \textbf{Estimation of Numeral Magnitudes (I)}: Within our models, the first notable phenomena we observe is that for the decoding and addition tasks designed to assess the fidelity of numerical magnitudes captured by the numeral embeddings, the logarithmic compression (\textit{ln} Anchors) has a greater contribution to the model performance than directional anchors (Anchors ($\footnotesize{\rightleftarrows}$)).
     \item \textbf{Estimation of Numeral Magnitudes (II)}: As the GMM-based anchors favor numerals frequent in the corpus, the anchors become sparse at higher numeral ranges - $[10k, 10^{10}]$. Thus, for this range specifically, we see that the model that strictly relies on directional anchors outperforms the log-compressive anchors on magnitude estimation tasks. Essentially, when the anchors are further from each other, knowing which direction they reside in with respect to the target numeral aids the model in reasoning about that numeral.
     \item \textbf{Estimation of Relative Ordering}: The second phenomena we observe is that for the task of retrieving the maximum/minimum numeral from a list of numerals, designed to assess the relative ordering capabilities of the numeral embeddings, the model that leverages both compressive representations and directional priming \cite{reynvoet:02} (\textit{ln} Anchors ($\footnotesize{\rightleftarrows}$)), has the best performance. Establishing that the incorporation of directional priming through the use of directional anchors further increases the relative ordering capabilities of the numeral embeddings.     
 \end{itemize}

 For easier comparisons among models, the measure employed for the decoding and addition tasks is \textit{log-RMSE}; as the error is log-compressed, seemingly small changes to the log-RMSE score translates to visible changes in numerical estimation through their embeddings, as depicted in Figure \ref{fig:plots}.

\section{Conclusions}
In this paper, we have presented a simple
plug-and-play BERT variant with enhanced numerical capabilites. Through our rigorous interpolation (in-domain) and extrapolation (out-of-domain) analyses, we showcase the superiority of our model in numeric comprehension while outlining the impact of logarithmic compression on magnitude estimation and the impact of directionality on relative ordering capabilities. Further, as a consequence of introducing anchors, we find the learning of niche pockets of similar embeddings for numerals closer in their magnitudes (\textcolor{blue}{\textsection A.4}).

\section{Related Work}
Although the majority of recent scholarly work in this domain revolves around training models to solve math problems \cite{mathnlp:17, mathnlp:21, liang:22} or strict arithmetic \cite{sharma:2022_2, sharma:2023}, several notable articles have looked exclusively into numeracy. \citet{spith:18} and \citet{jiang:20} devise strategies with Gaussian mixture models to generate embeddings for out-of-vocabulary numeral tokens. Similarly, \citet{razeghi:22} study the impact of numeral frequency in the pre-training corpus for few-shot arithmetic reasoning. \citet{naik:18}, \citet{wallace:19}, and \citet{pal:21} perform exploratory analysis of numeric comprehension through probing strategies. 

\section*{Limitations}
The restrictions from our in-house GPU resources do not allow scaling this study to more recent models that exceed 1 billion parameters. Nevertheless, recently published baselines that we evaluate against use the same underlying architecture that we employ, viz. the base BERT model. Given that larger models also depend on the base transformer architecture \cite{attention:og} and use similar learning mechanisms, we believe that these observations will carry over to larger models as well.

\section*{Ethics Statement}
The datasets we use in this study are established benchmark datasets from publicly accessible websites and do not contain any personally identifiable information. Our analyses does not constitute human subjects and thus do not fall within the purview of the IRB.

\bibliography{custom}
\bibliographystyle{acl_natbib}

\appendix

\section{Appendix}
\label{sec:appendix}
\begin{table*}[t]
\centering
\scriptsize
\caption{\footnotesize \textit{Anchors generalize much better to unseen OOD numerals:} Performance of our model variants (Anchors) vs the baselines for out-of-domain numerals on four tasks evaluating the numeracy captured by model embeddings. The tasks are further sub-divided into number ranges and column $\forall$ Z $\in$ C includes all numerals $Z$ in corpus $C$.}
\begin{tabular}{@{}lcccc@{}}
\toprule
Models     & \multicolumn{2}{c}{\textbf{Decoding} (Log-RMSE)}     & \multicolumn{2}{c}{\textbf{Addition} (Log-RMSE)}     \\ \midrule
Range      & OOD {[}1k, 10k{]}   & OOD {[}10k, $10^{10}${]}  & OOD {[}1k, 10k{]}   & OOD {[}10k, $10^{10}${]}  \\ \midrule
GenBERT    & 0.0132              & 0.0602                & 0.0130              & 0.0922                \\
MWP-BERT   & 0.0097              & 0.0537                & 0.1205              & 0.0788                \\ \midrule
Anchors        & 0.0059              & 0.0328                & 0.0082              & 0.0419                \\
Anchors (\scriptsize{$\rightleftarrows$})     & 0.0043              & {\ul 0.0278}          & 0.0067              & {\ul 0.0409}          \\
\textit{ln} Anchors    & 0.0033              & 0.0338                & 0.0043              & 0.0557                \\
\textit{ln} Anchors (\scriptsize{$\rightleftarrows$}) & {\ul 0.0029}        & 0.0347                & {\ul 0.0033}        & 0.0625                \\ \midrule
           & \multicolumn{2}{c}{\textbf{List Maximum} (Accuracy)} & \multicolumn{2}{c}{\textbf{List Minimum} (Accuracy)} \\ \midrule
           & OOD {[}1k, 10k{]}   & OOD {[}10k, $10^{10}${]}  & OOD {[}1k, 10k{]}   & OOD {[}10k, $10^{10}${]}  \\ \midrule
GenBERT    & 86.50\%             & 78.49\%               & 90.00\%             & 76.00\%               \\
MWP-BERT   & 87.00\%             & 82.50\%               & 88.50\%             & 77.00\%               \\ \midrule
Anchors        & 84.50\%             & 83.50\%               & 89.49\%             & 83.50\%               \\
Anchors (\scriptsize{$\rightleftarrows$})      & 86.00\%             & {\ul 88.50\%}         & 90.00\%             & 81.50\%               \\
\textit{ln} Anchors    & 87.50\%             & 86.99\%               & 90.00\%             & 83.50\%               \\
\textit{ln} Anchors (\scriptsize{$\rightleftarrows$}) & {\ul 88.00\%}       & 87.00\%               & {\ul 91.50\%}       & {\ul 84.00\%}         \\ \bottomrule
\end{tabular}
\label{table:2}
\end{table*}

\subsection{Gaussian Mixture Models Initialization and Parameters}
As Gaussian mixture models are sensitive to initialization methods \cite{blomer:13}, we initialize our models with random sampling from the dataset. The heterogeneous nature of the numeral distribution in the dataset lends this as the optimal initialization strategy. The models are trained to a convergence tolerance of 0.001 with each component given its own general covariance matrix. The choice of $K$ = 1000 Gaussian components was established stabilizing AIC and BIC values through a parameter sweep with $K$ ranging from 10 to 5000.

\subsection{Experimental Setup}
\subsubsection{Training Corpus}
The WikiText-103 corpus \cite{wikitext} consists of  611,725 training instances (that includes over 100 million tokens) extracted from the set of verified \textit{good} and \textit{featured} articles on Wikipedia. Numeral tokens account for 2.4\% of the corpus tokens with quadruple-digit numbers accounting for the greatest concentration of numerals  - 41.8\% .

\subsubsection{Training Configurations}
For both our baselines GenBERT \cite{mathnlp:20} and MWP-BERT \cite{liang:22}, the pre-trained models that the authors have provided are used as-is, thus ensuring no performance degradation as a consequence of in-house training/replication. For our Anchor models, the scheme for training follows BERT's standard training protocol of using masked-language modeling. However, instead of randomly masking 15\% of the tokens as done in BERT, we mask the anchor numeral as we intend to ground the learning of the target numerals based on their anchors. With the standard sequence size of 512 for BERT, the models were trained for 6 epochs each in a cluster of 4 Tesla P100 GPUs. The pre-trained BERT models are loaded from the Huggingface library \citep{huggingface}.

\subsubsection{Embedding Retrieval}
As recommended in the original BERT configuration, we tested hidden representations from the last hidden layer as well as from the sum of the last 4 hidden layers. We observed the best performance using a sum of the last 4 hidden layer representations, which we adopt for our experimentation.

\subsubsection{Regressors and Classifiers}
For consistency in our experimental results, we opted for Extreme Gradient Boosting (XGBoost) \cite{xgboost} for regression over standard neural networks for their robustness to parameterization. The regressors were initialized with 1000 components with each tree having a maximum depth of 5 and trained with a learning rate of 0.01. Similarly, a standard LSTM setup with 4 stacked LSTMs coupled with a sigmoid activation for the final linear layer was used as the classifier. Each classifier was trained for 150 epochs with a learning rate of 1e-4.

\subsection{Extrapolation for Out-domain Numerals}
As depicted in Table \ref{table:1} for in-domain numerals, we perform the same set of evaluations for out-of-domain (unseen) numerals in Table \ref{table:2}, corroborating the same performance gains that we observed for in-domain numerals. Please note that all numerals in range [1,100] and [100, 1k] appear in the training corpus, thus only the ranges [1k, 10k] and [10k, $10^{10}$] qualify for OOD evaluation. 

\subsection{Embedding Visualizations}
\begin{figure*}[b]
    \centering
    \subfloat[\footnotesize \centering Base BERT model]
    {{\includegraphics[scale=0.35]{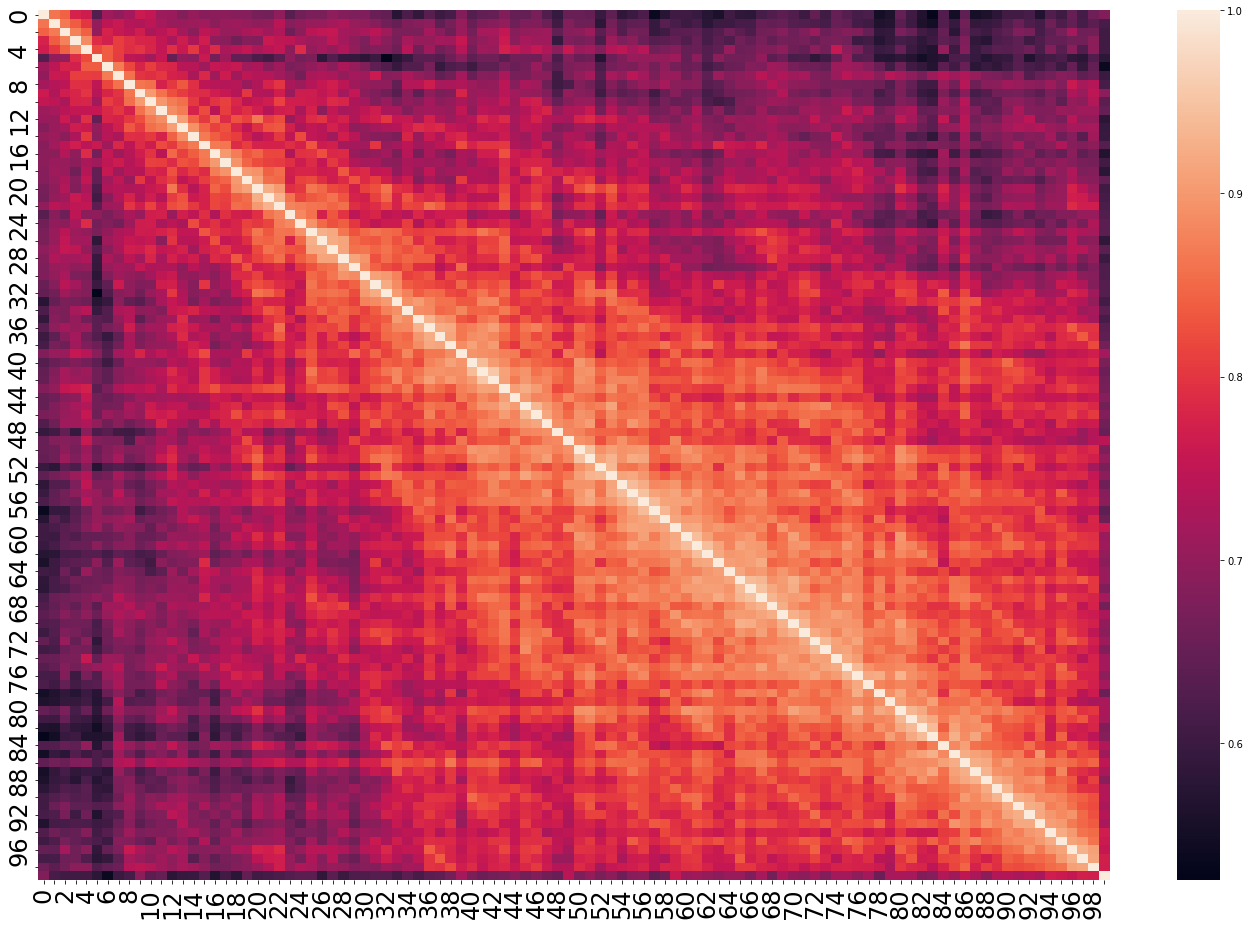} }} \\
    \subfloat[\footnotesize \centering Our model]{{\includegraphics[scale=0.35]{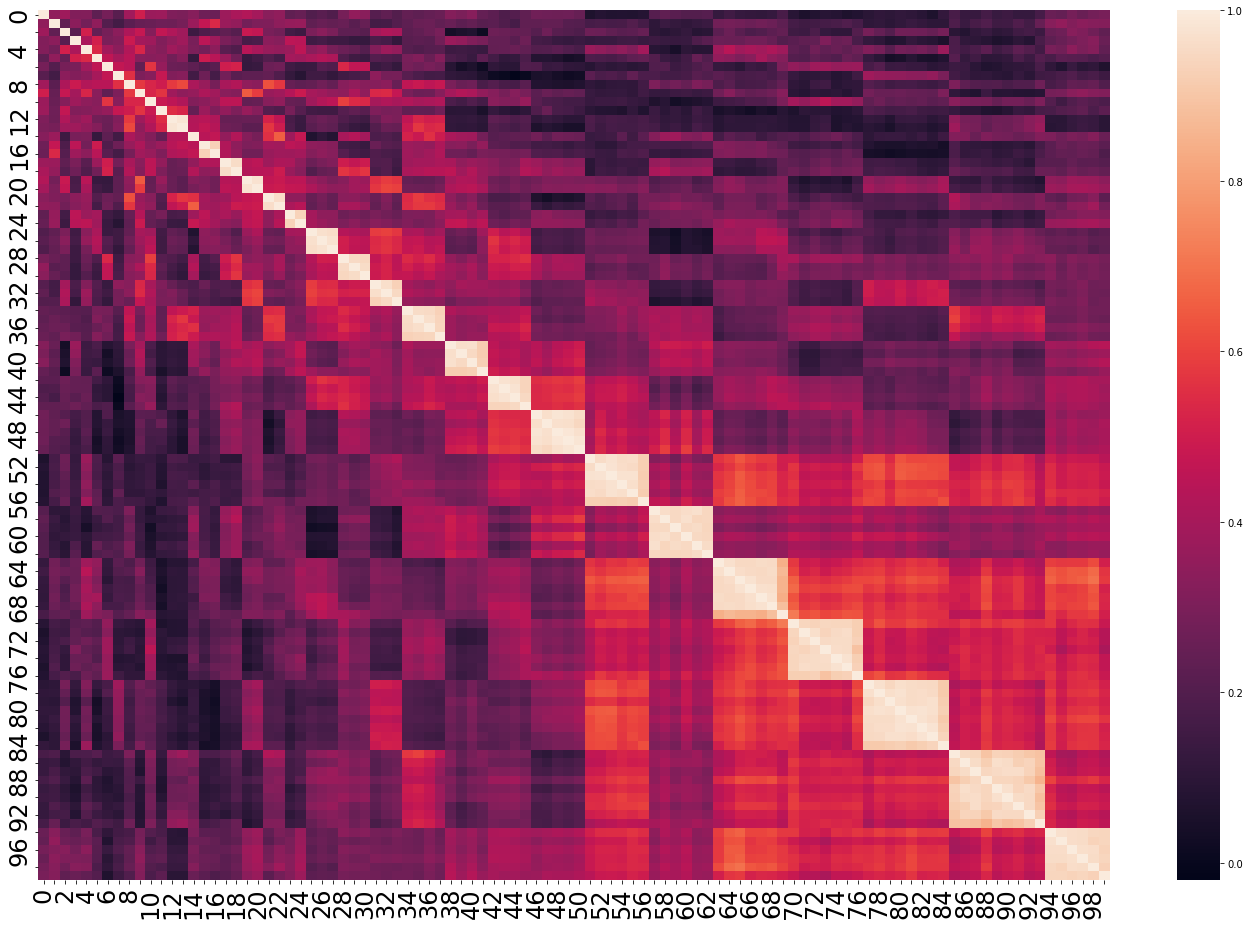} }}%
    \caption{\footnotesize Heatmaps computed from cosine similarities of numeral embeddings in range [1,100].}%
    \label{fig:heatmap}
\end{figure*}

As an alternative visualization tool, we contrast heatmaps generated through the cosine similarities of numeral embeddings for the base BERT model and our model. As illustrated in Figure \ref{fig:heatmap}, the heatmap for the base BERT model has uniformly low cosine similarity throughout, leading to little distinction between numeral embeddings. In contrast, the heatmap for our model demonstrates sophisticated patterns of similarity for proximal numerals along its diagonal. Also seen are sections of low similarity scores in the top right and bottom left - indicating the ability to discern numerical magnitudes of lower and higher number ranges.

\end{document}